\documentclass{article}
\usepackage{spconf,amsmath,graphicx}
\usepackage{xspace}
\usepackage{adjustbox}
\usepackage{xcolor}
\definecolor{mywhite}{rgb}{1, 1, 1}
\usepackage{amsmath}
\usepackage{cite}
\usepackage{amsmath,amssymb,amsfonts}
\usepackage{algorithmic}
\usepackage{graphicx}
\usepackage{textcomp}
\usepackage{color}
\usepackage{algorithm}
\usepackage{algorithmic}
\usepackage{lipsum}
\usepackage{mathtools}
\usepackage{amsmath}
\usepackage{dsfont}
\usepackage{bbm}
\usepackage{hyperref}

\usepackage{subfigure}
\usepackage{caption}

\DeclareMathOperator{\Lagr}{\mathcal{L}}

\newcommand\blfootnote[1]{%
  \begingroup
  \renewcommand\thefootnote{}\footnote{#1}%
  \addtocounter{footnote}{-1}%
  \endgroup
}

\title{SVBR-Net: A Non-Blind Spatially Varying Defocus Blur Removal Network}

\name{Ali Karaali$^{1,2}$ \qquad Cl\'{a}udio Rosito Jung$^{3}$ }
\address{
$^{1}$ Department of Medical Gerontology, Trinity College Dublin, Ireland \\
$^{2}$ ADAPT: SFI Centre for Digital Media Technology, Ireland\\
$^{3}$ Institute of Informatics, Federal University of Rio Grande Do Sul, Porto Alegre, RS, Brazil \\
karaalia@tcd.ie, crjung@inf.ufrgs.br
}
\begin{document}
%
\maketitle

\begin{abstract}
\vskip -0.05cm
Defocus blur is a physical consequence of the optical sensors used in most cameras. Although it can be used as a photographic style, it is commonly viewed as an image degradation modeled as the convolution of a sharp image with a spatially-varying blur kernel. Motivated by the advance of blur estimation methods in the past years, we propose a non-blind approach for image deblurring that can deal with spatially-varying kernels. We introduce two encoder-decoder sub-networks that are fed with the blurry image and the estimated blur map, respectively, and produce as output the deblurred (deconvolved) image. Each sub-network presents several skip connections that allow data propagation from layers spread apart, and also inter-subnetwork skip connections that ease the communication between the modules. The network is trained with synthetically blur kernels that are augmented to emulate blur maps produced by existing blur estimation methods, and our experimental results show that our method works well when combined with a variety of blur estimation methods. Our code will be available at \url{https://github.com/alikaraali/SVBR_Net_icip                                                                                                                                                                                                                                                                                                                                                                                                                }.
\end{abstract}
\vskip -0.15cm

\vskip -0.75cm
\section{Introduction}
\label{sec:intro}
\vskip -0.25cm

Defocus blur\blfootnote{``\copyright 2022 IEEE. Personal use of this material is permitted. Permission from IEEE must be obtained for all other uses, in any current or future media, including reprinting/republishing this material for advertising or promotional purposes, creating new collective works, for resale or redistribution to servers or lists, or reuse of any copyrighted component of this work in other works.''} is mostly noticeable in shallow DoF cameras, and its amount is spatially variant: objects farther from the focal plane project to blurrier regions. We can characterize the blur amount by assigning a ``spread level'' parameter for each pixel, which is often represented by a disk radius $C_{\sigma_b}$ or a Gaussian variance $\sigma^2_b$. 
Mathematically, the captured blurry image $I_b$ can be formulated as the convolution of the in-focus image $I$ and the spatially varying point-spread function (PSF) $h_{\sigma}$ plus noise $\eta$, given by
\begin{equation}
I_b(x,y) = (I * h_{\sigma(x,y)}) (x,y) + \eta(x,y),
\label{eq:blur}
\end{equation}
where $(x,y)$ is the spatial position of a pixel.

Deblurring consists of recovering the original sharp image $I$ from a blurry image $I_b$, which involves undoing the convolution (deconvolution). Deconvolution methods can be classified into two main classes: blind and non-blind. Blind approaches work without knowing the degradation kernel to estimate the original image (the kernels might be estimated in the process for some methods). Non-blind approaches assume that the convolutional kernel is known, and focus on the restoration process itself. Either blind or non-blind deconvolutions are highly ill-posed~\cite{levin1}, especially for spatially varying kernels, which makes image deblurring a challenging task.

\begin{figure*}[!t]
    \begin{center}

\subfigure[Overview of the proposed network for non-blind spatially varying defocus blur removal. Each color represent different type of neural block, and arrow represents the post concatenation. Each block also shows the dimension of the output feature.]{
    \includegraphics[width=0.9\textwidth]{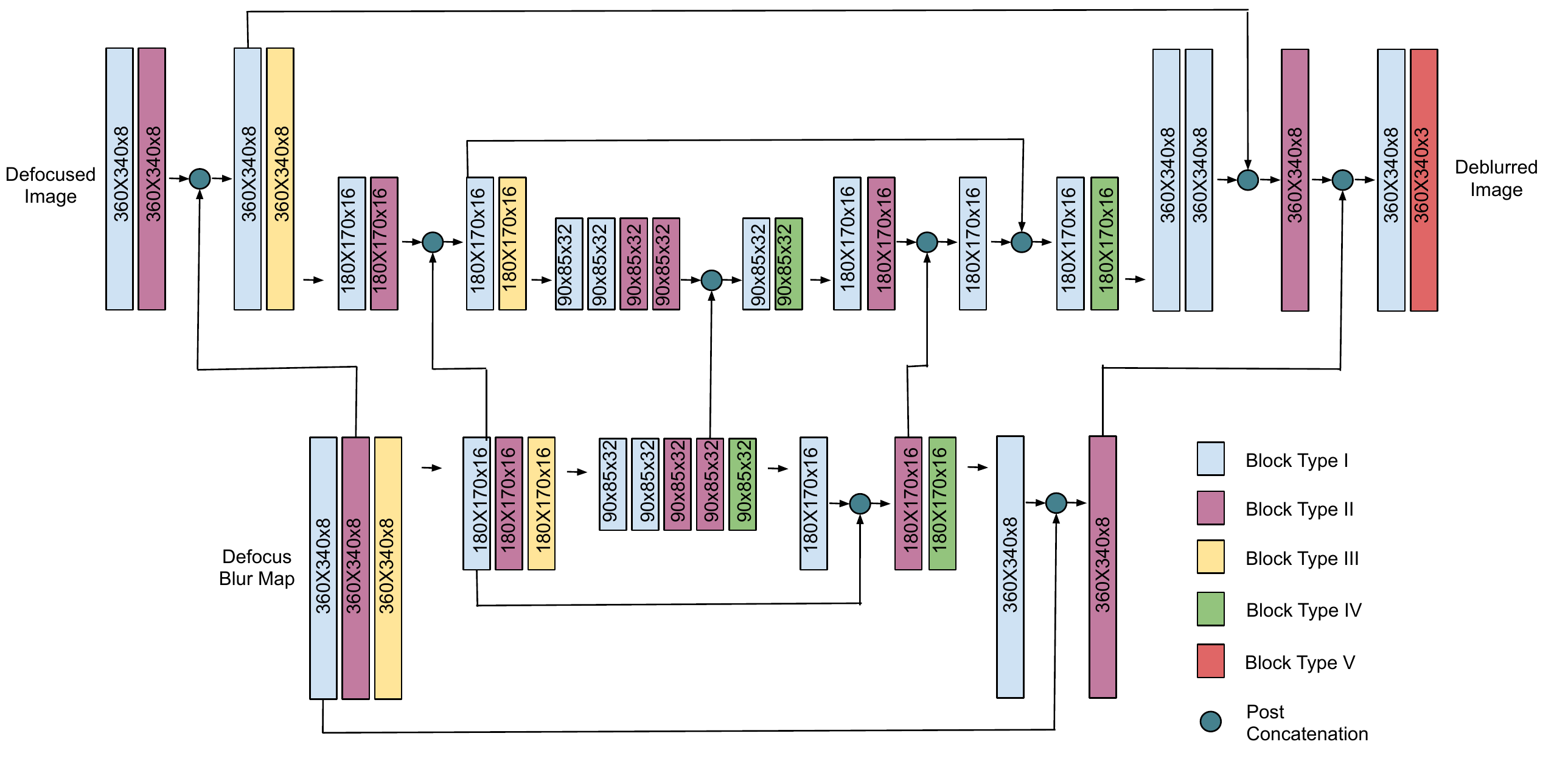}
    \label{fig:networkdesig}
}
    \vskip -0.5cm
\subfigure[Overview of the neural block: \textit{Block Type I}.]{
    \includegraphics[width=.3\linewidth]{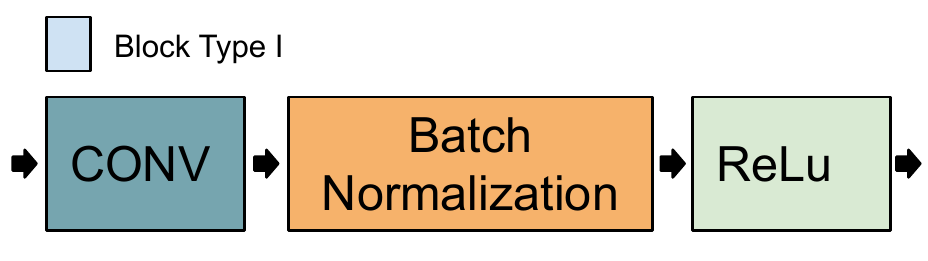}
    \label{fig:block1} 
}
\subfigure[Overview of the neural block: \textit{Block Type II}.]{
    \includegraphics[width=.3\linewidth]{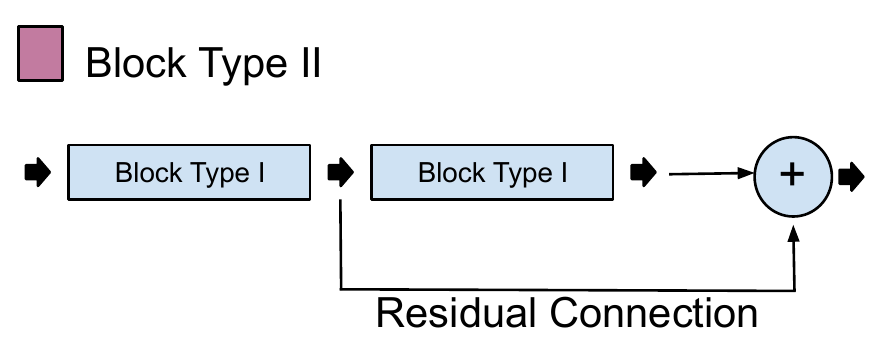}
    \label{fig:block2}
}
\subfigure[Overview of the neural block: \textit{Block Type III}.]{
    \includegraphics[width=.3\linewidth]{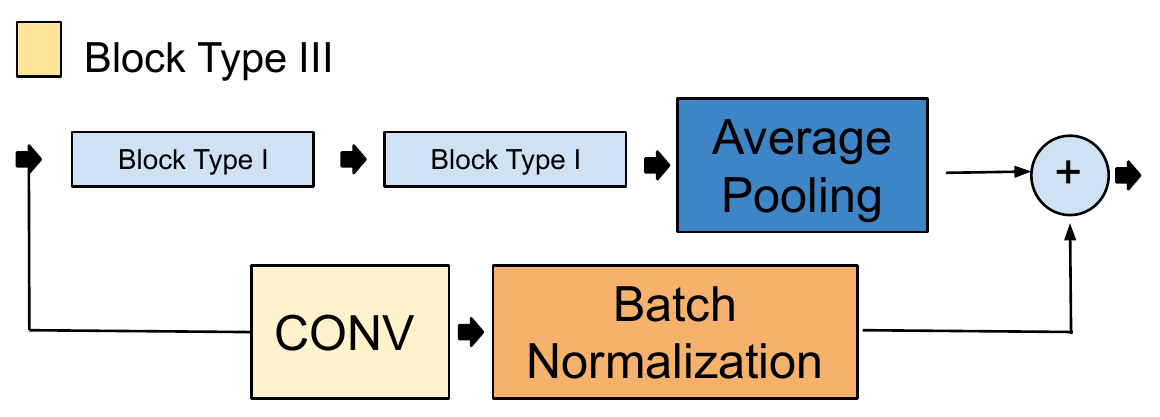}
    \label{fig:block3}
}
\vskip -0.5cm
\subfigure[Overview of the neural block: \textit{Block Type IV}.]{
    \includegraphics[width=.4\linewidth]{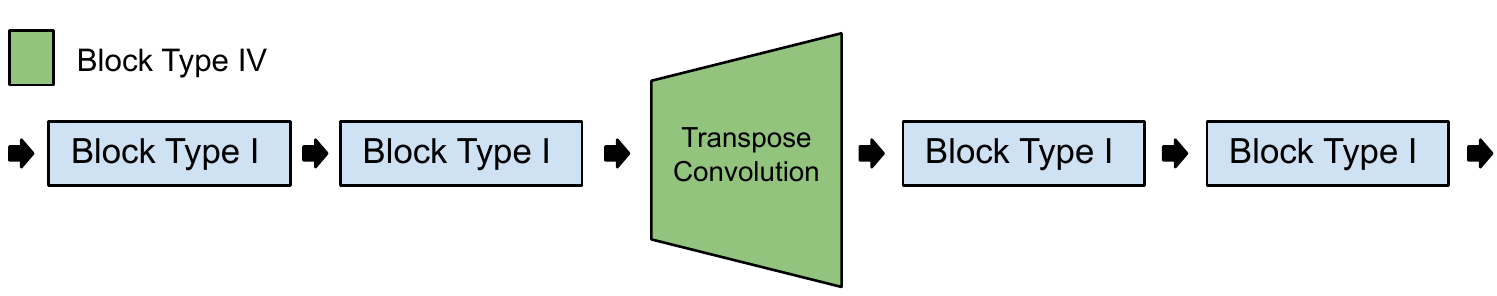}
    \label{fig:block4}
}
\subfigure[Overview of the neural block: \textit{Block Type V}.]{
    \includegraphics[width=.2\linewidth]{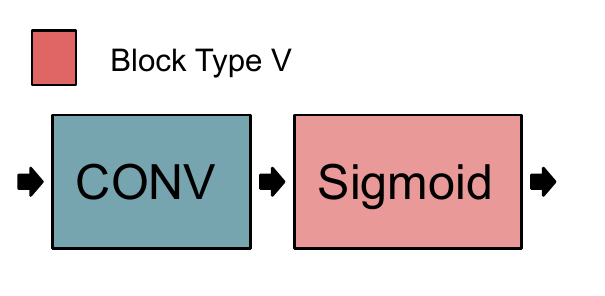}
    \label{fig:block5}
}
\vskip -0.5cm
\caption{Overall architecture of the proposed SVBR-Net: A Non-Blind Spatially Varying Defocus Blur Removal Network} \label{fig:overview}
\end{center}
\vskip -1.0cm
\end{figure*}

\noindent \textbf{Blind methods} aim to estimate the input image $I$ and the kernel $h$ (and sometimes the noise estimate $\eta(x,y)$) such that Eq.~\eqref{eq:blur} is satisfied according to an error metric (e.g., MSE).
Since this is an ill-posed problem, additional constraints must be provided, such as $H^1$ regularization terms
~\cite{you1996regularization} or using total variation (TV) regularization~\cite{chan1998total}. Motivated by the introduction of datasets and challenges (such as NTIRE~\cite{nah2021ntire}), several deep learning approaches have been developed for blind image deblurring (focusing mostly on motion blur). Xu and collaborators~\cite{xu2014deep} proposed a deep convolutional network to approximate the pseudo inverse kernel for deconvolution applications. Tao et al.~\cite{tao2018scale} presented a Scale-Recurrent Network (SRN-DeblurNet) that performs image deblurring in a coarse-to-fine fashion, focusing their analysis no motion blur. Kupyin and collaborators~\cite{kupyn2019deblurgan} proposed an end-to-end generative adversarial network (GAN) for motion image deblurring. Ma et al.~\cite{9619959} recently proposed a defocus deblurring approach that learns the spatially varying blur map as an auxiliary task. Despite the promising deblurring results, the produced blur maps are still very noisy.

In a highly related problem, defocus blur estimation approaches aim to estimate the spatially-varying blur kernel $h(x,y)$ from a single blurry image~\cite{TIPpaper20161,DefocusPaper,karaali2017edge,pattern2020,domainadapt,tip2022}, with increasingly better results. Such advance might leverage the development of \textbf{non-blind deblurring methods}, which assume that both $I_b$ and $h$ are known. Despite being a theoretically simpler problem when compared to the blind version, it is still an ill-posed problem.  
A broad analysis on Bayesian image restoration was presented by Geman and Geman~\cite{geman1984stochastic}, with the foundations for several statistical tools to handle the problem. The Maximum a Posteriori (MAP) framework can also be adapted to the context of non-blind deconvolution by dropping $h$ as an unknown and imposing priors only on the image domain, as explored by Levin et al.~\cite{levin2007image}. Dabov and colleagues~\cite{dabov2008image} presented a deconvolution method suited for spatially-invariant kernels using regularization by block-matching and 3D (BM3D) filtering, whereas Graham and Yithzaky~\cite{graham2019blind} presented a deblurring approach in which the final image is obtained by several layers with different blur amounts.

A common drawback of non-blind approaches is the coupling between the blur estimation methods and the restoration process: bad blur estimates tend to produce poor-quality deblurred images. This work presents a novel end-to-end deep neural network (DNN) for non-blind image deblurring that takes as input a dense blur map and the blurry image $I_b$, and outputs a deblurred version of $I_b$. Strong data augmentation is applied to make the network robust to a variety of errors and artifacts that are present in blur estimation methods, which are used to produce the required blur maps.

\section{The Proposed Method}
\vskip -0.25cm

\noindent
\textbf{Network architecture:} the proposed network, called SVBR-Net (Spatially Varying Blur Removal Network), is based on a pair of U-Net encoder-decoders~\cite{unetpaper} that deal with the two inputs (the normalized blurry image and defocus map, both with spatial resolution $H \times W$). Each branch presents skip connections (intra-subnet) to facilitate data propagation through the network (across multiple resolutions), and the two branches present several interconnections (inter-subnet) that leverage the propagation of information across blocks from the two networks with compatible dimensions, as illustrated in Fig.~\ref{fig:overview}. 
Each branch of the encoder-decoder is composed by five different types of neural blocks (visual definitions in Figs.~\ref{fig:block1}-\ref{fig:block5}). Type I is composed basically by a conv-batch module with a $3\times 3$ kernel and a variable number of filters (specified in Fig.~\ref{fig:networkdesig}), and type II is a composition of two type I blocks with residual connections. Type III blocks are built by summing the output of two parallel branches: i) two consecutive type I blocks followed by average pooling layer, and ii) $2 \times 2$ unpadded convolution with $2$ pixel followed by a batch normalization layer. Type III blocks double the number of channels and halve the spatial resolution. Type IV is used in the decoder, and it unwraps the coded deep features in the previous layer to a shallower but with better spatial resolution representation using transpose convolutions. Finally, type V is the output block that applies the final convolution operation to the information gathered by the two sub-networks and produces a normalized RGB image.

\begin{figure}[!t]
    \begin{center}
        \includegraphics[width=0.21\columnwidth]{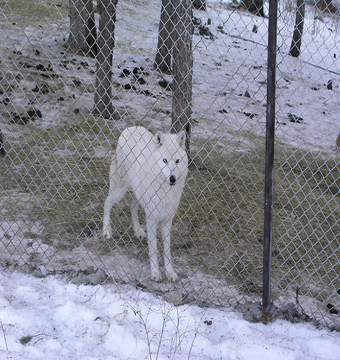}
        \includegraphics[width=0.21\columnwidth]{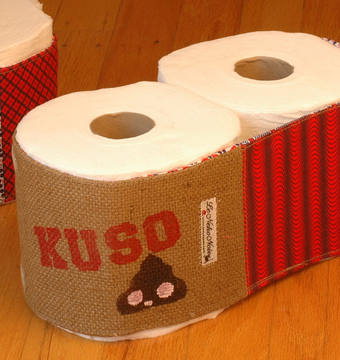}
        \includegraphics[width=0.21\columnwidth]{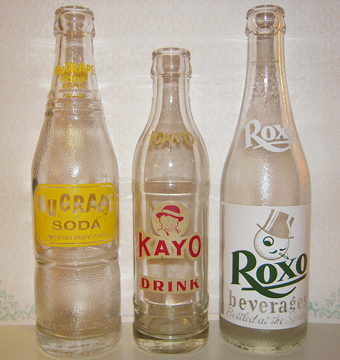}
        \includegraphics[width=0.21\columnwidth]{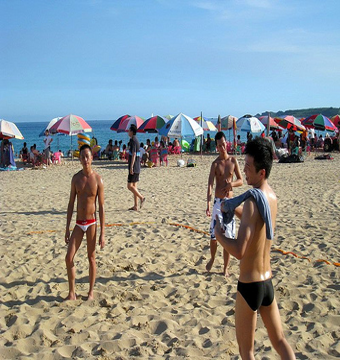} \\
    
        \includegraphics[width=0.21\columnwidth]{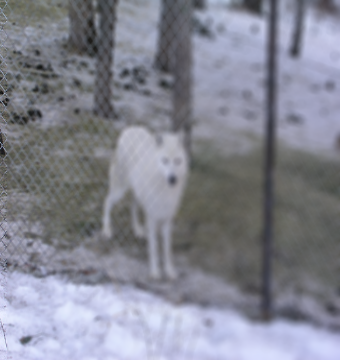}
        \includegraphics[width=0.21\columnwidth]{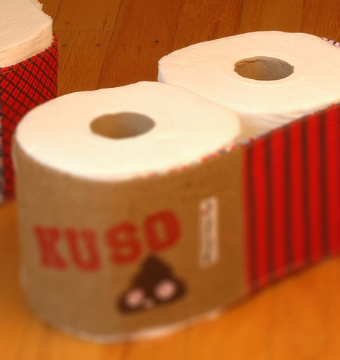}
        \includegraphics[width=0.21\columnwidth]{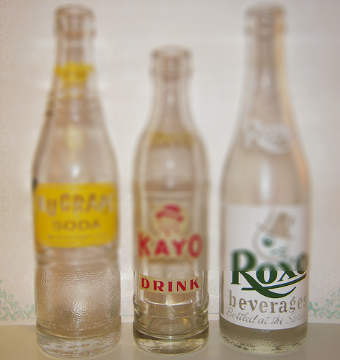}
        \includegraphics[width=0.21\columnwidth]{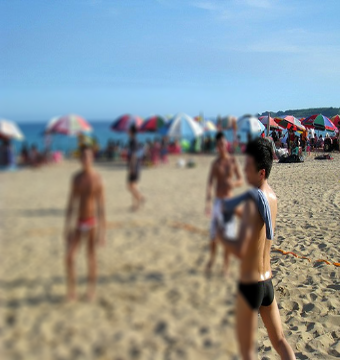} \\

        \includegraphics[width=0.21\columnwidth]{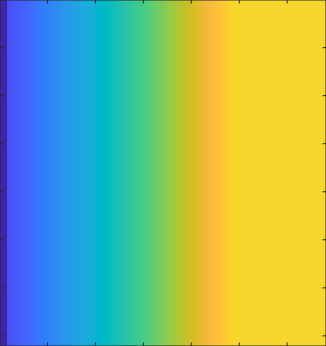}
        \includegraphics[width=0.21\columnwidth]{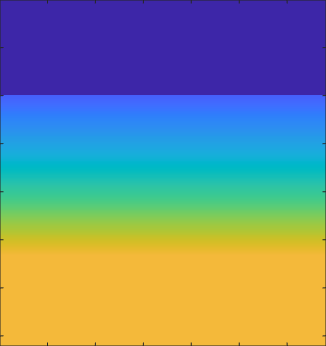}
        \includegraphics[width=0.21\columnwidth]{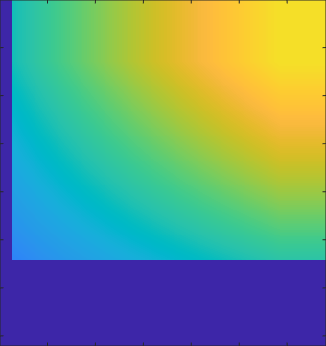}
        \includegraphics[width=0.21\columnwidth]{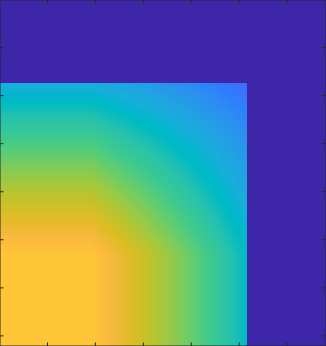} \\

        \includegraphics[width=0.21\columnwidth]{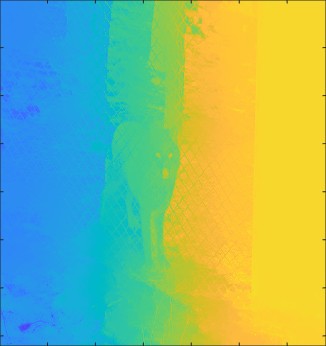}
        \includegraphics[width=0.21\columnwidth]{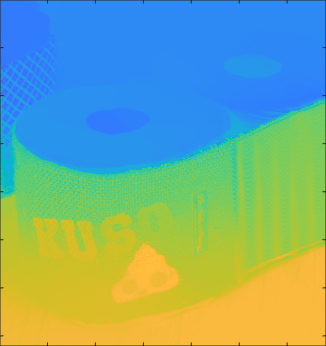}
        \includegraphics[width=0.21\columnwidth]{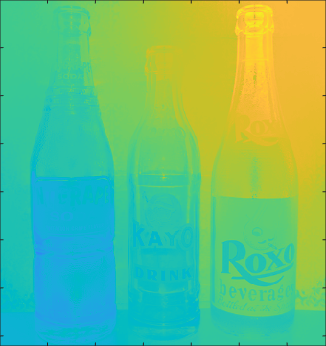}
        \includegraphics[width=0.21\columnwidth]{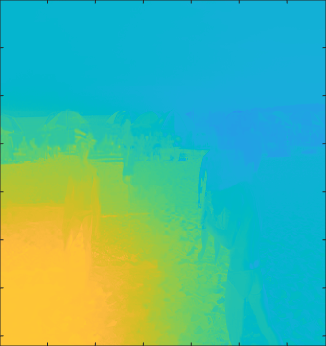} \\

        \includegraphics[width=0.21\columnwidth]{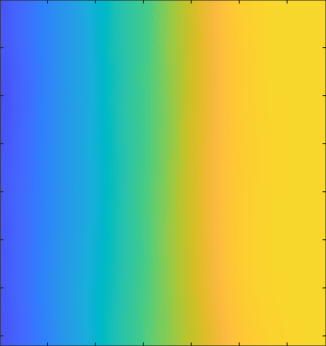}
        \includegraphics[width=0.21\columnwidth]{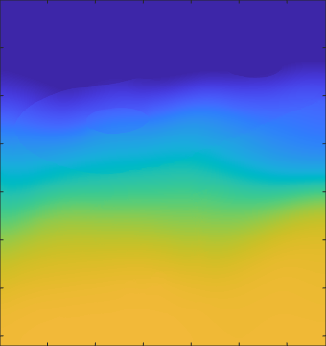}
        \includegraphics[width=0.21\columnwidth]{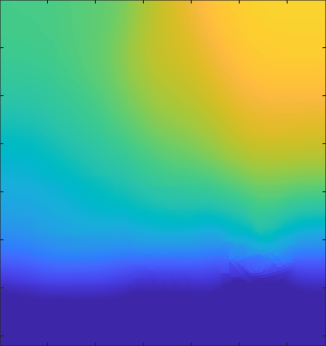}
        \includegraphics[width=0.21\columnwidth]{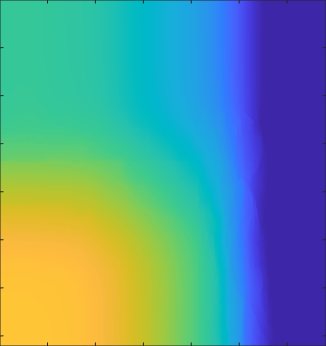} \\
        \includegraphics[width=0.5\columnwidth]{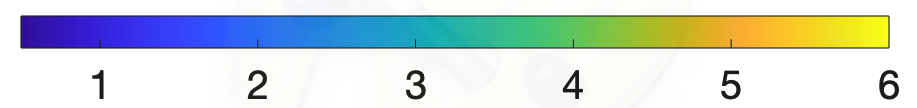} 
        \vskip -0.25cm
        \caption{\label{fig:syntheticdefocus} All-in-focus images (first row) and corresponding synthetically defocused images (second row). Remaining rows: true PSFs (third row), and interpolated PSFs by alpha-Laplacian Matting~\cite{LaplacianM} (forth row) and Domain Transform (DT) filtering~\cite{Gastal} (fifth row).}

    \end{center}
    \vskip -0.75cm
\end{figure}

\noindent \textbf{Data Preparation:} Due to lack of publicly available data, we created our own training set using synthetic blur kernels. We first carefully chose focused images (with no visually apparent blur) from two publicly available datasets: 350 images from ILSVRC and 350 images MS-COCO. The selected images are then re-scaled to $H \times W$, and convolved with several spatially-varying point spread functions (PSFs), which mimic defocus blur. More precisely, we created $N$ different PSFs  $h_{\sigma_i}$ where $i=1,2,...N$ that simulate a variety of different real-world depth change patterns (see third row of Fig.\ref{fig:syntheticdefocus}). Following the work of D'Andres~\cite{TIPpaper20161}, we utilize $23$ disk blur scales, starting from $C_{\sigma_1} = 0.5$ and increasing up to $C_{\sigma_{23}} = 6$\footnote{Blur level $C_{\sigma}=6$ as disk radii is able to cover the max defocus blur level for all image resolution levels~\cite{pattern2020}.} with a step size of $0.25$ to create the PSFs. We empirically set $N=39$ spatially-varying PSFs.

Since inference is performed using estimated blur maps (and not the actual blur values), training SVBR-Net with ground-truth (GT) blur maps only would lead to overfitting and compromise the deconvolution process when a noisy blur estimate is used. To mitigate this problem, we created a second set of PSFs to train the model, aiming to add robustness to the network. Our augmentation strategy was inspired by edge-based defocus blur estimation methods ~\cite{DefocusPaper,TIPpaper20161,karaali2017edge}, which first estimate the blur amount at image edges and then propagate the sparse blur map to the full image. These methods tend to achieve good accuracy for the sparse map (since the blur is better noticed and captured at high-frequency structures of the image), but the propagation scheme tends to generate artifacts. Based on these observations, we generate ``alternative'' versions of the GT PSFs by retrieving the original blur values only at edge locations and propagating the blur map to the remaining pixels using interpolation strategies. We used two main schemes: alpha-Laplacian Matting~\cite{LaplacianM}  and Domain Transform (DT) filtering~\cite{Gastal}. Fig.\ref{fig:syntheticdefocus} shows examples of augmented blur maps obtained with these two interpolation schemes.

\noindent \textbf{Training:} we use our synthetic defocused images and their corresponding annotations (both original GT PSFs and interpolated PSFs) in the training process, using $0.8 / 0.2$ split ratio for training and validation. Adam optimizer is used with a batch size $8$. The initial learning rate is set to $10^{-3}$, which is divided by $10$ at every 20 epochs. The network is initially trained using the true PSFs as blur level annotations for 32 epochs (to ease convergence) and then trained with additional 32 epochs using the interpolated PSFs as blur level annotations. To enforce visual similarity between the deblurred image $\hat{I}$ and the sharp version $I$,
we used the per-channel average of the Structural Similarity Index (SSIM) as the loss function, given by
\begin{equation}
    \Lagr = \frac{1}{M} \sum_{m=1}^{M} \left(1 - \frac{1}{3} \sum_{c=1}^{3} \text{SSIM}(\hat{I_c}, I_c) \right),
\label{eq:lossfunction}
\end{equation}
 where $I_c$ and $\hat{I_c}$ are the $c^{th}$ color channels of the sharp and deconvolved images, respectively, and  $M$ is the number of training samples in a batch. 

\section{Experimental Results}
\vskip -0.25cm

To validate our model, we used the dataset provided in ~\cite{TIPpaper20161} that provides $22$ pairs of naturally defocused images and their corresponding all-in-focus versions. Although the dataset also provides the GT blur maps, we ran SVBR-Net with estimated blur maps using five different existing approaches: classical edge-based methods~\cite{DefocusPaper,karaali2017edge}, region-based methods~\cite{TIPpaper20161,pattern2020}, and a SOTA deep learning-based method~\cite{domainadapt}. The quantitative validation of the proposed method is carried out using the two traditional image quality measures: PSNR and SSIM. 

We compare the quality scores of the deblurred images $\hat{I}$ with the results of other two spatially varying defocus blur removal methods: a combination of the deconvolution methods proposed in~\cite{comb1} and~\cite{comb2}, which is also used in \cite{TIPpaper20161,karaali2017edge,pattern2020,tip2022}, and the method proposed in~\cite{zhang2016spatially}. Additionally, we compare the proposed method with a very recent blind approach~\cite{9619959} that estimates both the blur map and the deblurred image.

\begin{table}[t]
\caption{\label{tab:table1} Mean SSIM/PSNR values of the deblurred images using the true PSF and estimated PSFs produced the existing methods proposed in \cite{TIPpaper20161,DefocusPaper,karaali2017edge,pattern2020,domainadapt}), along with MAE (mean absolute error) of estimated blur maps.}
\vskip -.35cm
\centering
\scalebox{0.55}{
\begin{tabular} {c c c c c c c} 
 \hline
& GT Map & \cite{TIPpaper20161} & \cite{DefocusPaper} & \cite{karaali2017edge} & \cite{pattern2020} & \cite{domainadapt}\\ [0.5ex] 
 \hline\hline
MAE                            & N/A   & 0.197 &   0.589 &    0.378 &    0.172 & 0.565 \\
 \hline
\cite{zhang2016spatially}      & 0.756/22.06          & 0.776/23.39                   & 0.725/20.99                   & 0.756/21.71                   & 0.754/21.63          & 0.781/23.68 \\
~\cite{comb1} \& ~\cite{comb2} & 0.885/\textbf{27.69} & 0.862/26.45                   & 0.788/23.83                   & 0.833/25.72                   & 0.865/\textbf{26.61} & 0.803/24.65  \\
SVBR-Net GTB.               & 0.900/26.45          & 0.876/25.47                   & 0.732/22.23                   &   0.824/24.62                 &   0.881/25.79        &   0.765/23.49 \\ 
SVBR-Net                       & \textbf{0.902}/26.62 & \textbf{0.901}/\textbf{26.59} & \textbf{0.894}/\textbf{26.41} & \textbf{0.898}/\textbf{26.52} & \textbf{0.901}/26.57 & \textbf{0.897}/\textbf{26.43} \\ 
 \hline
 \cite{9619959} & \multicolumn{6}{c}{0.855/26.08} \\
 \hline
\end{tabular}
}
\vskip -0.75cm
\end{table}

Table~\ref{tab:table1} shows the SSIM and PSNR quality scores of the proposed approach and other competitive deblurring methods. We also show the accuracy of the blur estimation methods used as input (in terms of the Mean Absolute Error -- MAE), noting that smaller MAE values mean better blur estimates. We observe that most of the deconvolution methods produce their best results when fed by the GT defocus blur map (GT Map), while the quality score fluctuates depending on the used estimated defocus blur map. The proposed network (SVBR-Net) also produced the best results when fed with the GT blur map, but the decrease in quality was very small compared to all the estimated blur maps. In particular, using the outputs of both \cite{TIPpaper20161} and \cite{pattern2020} lead to very similar results as those obtained with GT map in terms of SSIM and PSNR. Our quality indices for all tested blur maps were also higher than those reported in the recent blind deblurring method~\cite{9619959} (since it does not require blur estimates, only a single SSIM/PSNR pair is shown). To show the importance of the augmentation procedure, we also report the results training our method only using the ground-truth blur maps (called ``SVBR-Net GTB'' in Table~\ref{tab:table1}). We note that results using the GT blur maps are still good, but get highly degraded when estimated blur maps are employed.

\begin{figure}[!t]
    \begin{center}
        \includegraphics[width=0.32\columnwidth]{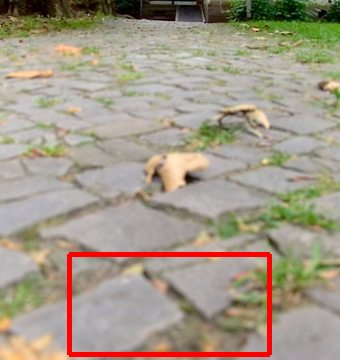}
        \includegraphics[width=0.32\columnwidth]{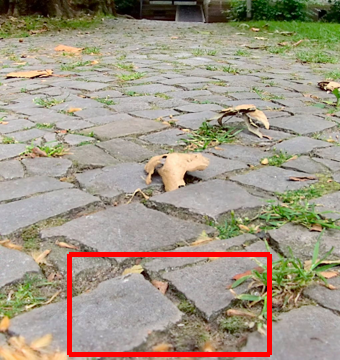}
        \includegraphics[width=0.32\columnwidth]{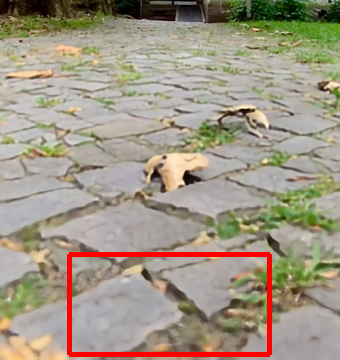}
        \\ \vskip .05cm
        \includegraphics[width=0.32\columnwidth]{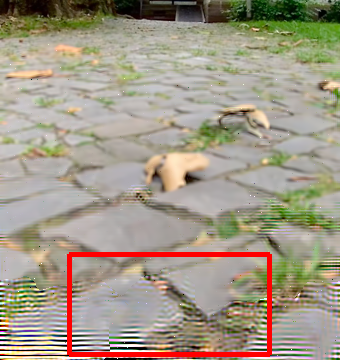}  
        \includegraphics[width=0.32\columnwidth]{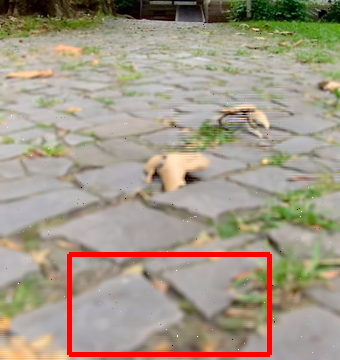}  
        \includegraphics[width=0.32\columnwidth]{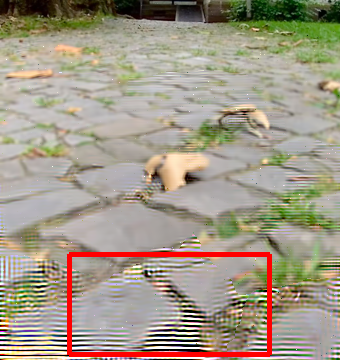}   \\ \vskip .05cm
        \includegraphics[width=0.32\columnwidth]{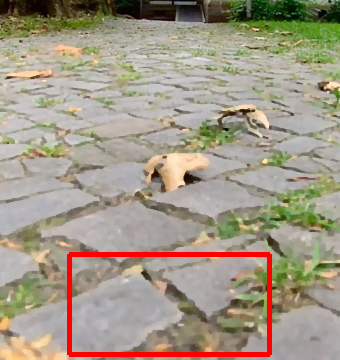}  
        \includegraphics[width=0.32\columnwidth]{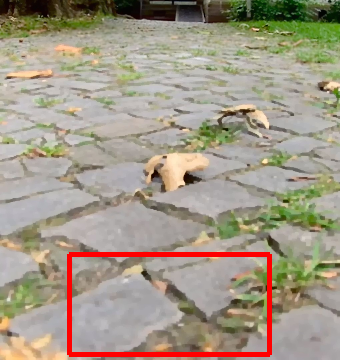}  
        \includegraphics[width=0.32\columnwidth]{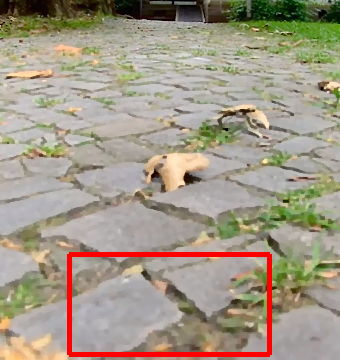}   \\ \vskip .05cm

        \includegraphics[width=0.32\columnwidth]{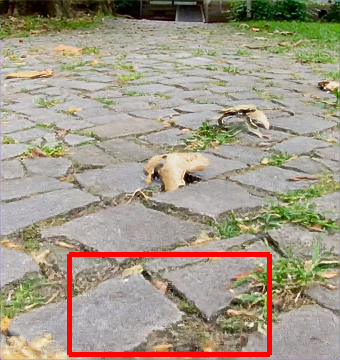}  
        \includegraphics[width=0.32\columnwidth]{r4c1_lab.png}  
        \includegraphics[width=0.32\columnwidth]{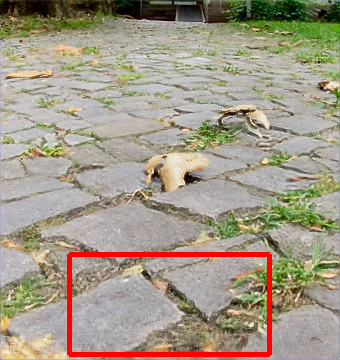}   \\
     \vskip -.25cm
        \caption{\label{fig:results1} Example of deblurring results. First row:  original defocused image, all-in-focus image (GT), and deblurred image using the blind approach~\cite{9619959}. Remaining rows show the results the competitive approaches, from top to bottom: \cite{zhang2016spatially}, a combination of~\cite{comb1} and ~\cite{comb2}, and the proposed SVBR-Net, respectively. Each column in rows 2--4 relates to a different blur map estimate, from left to right: ground-truth Map (GT Map), \cite{TIPpaper20161}, and \cite{pattern2020} from left to right, respectively. The red rectangles indicates blurrier regions in the original image.}
    \end{center}
     \vskip -0.75cm
\end{figure} 

Fig.~\ref{fig:results1} shows some deblurring examples from the provided database~\cite{TIPpaper20161}.
Although visual comparison might lead to a very subjective evaluation, we can observe that~\cite{zhang2016spatially} clearly produces artifacts (second row). On the other hand, the deblurring results combining~\cite{comb1} and~\cite{comb2} (third row) and the recent blind method proposed in \cite{9619959} (top right) does not manage to remove the blur sufficiently, particularly at the bottom of the images (highlighted with a red rectangle). On the other hand, the proposed network SVBR-Net produces focused images that are close to the original sharp image regardless of the employed blur estimation technique (three images in the last row).

\section{Conclusions}
\vskip -0.25cm

This paper presented a novel deep neural network for non-blind image deblurring tailored to spatially-variant blur, which is typically the case for defocus-related blur. We also propose a training protocol that requires only all-in-focus images and synthetically generated blur maps (with an augmentation procedure to emulate estimated blur maps), allowing a virtually limitless number of images.

The proposed approach was validated in a cross-dataset fashion: training was performed using images from ILSVRC and MS-COCO, and evaluated using the dataset proposed in~\cite{TIPpaper20161}. Our results indicate that our method presents better results than competitive approaches, particularly when lower-quality blur maps (e.g., produced by~\cite{DefocusPaper}) are used.

\section*{Acknowledgments}

This work was partly funded by the ADAPT Centre for Digital Content Technology, which is funded under the SFI Research Centres Programme (13/RC/2106\_P2) and is co-funded by the European Regional Development Fund, and also partly supported by Brazilian agencies CAPES and CNPq.



\vskip -0.25cm
\bibliographystyle{IEEEbib}
\bibliography{refs}

\begin{thebibliography}{10}

\bibitem{levin1}
A.~{Levin}, Y.~{Weiss}, F.~{Durand}, and W.~T. {Freeman},
\newblock ``Understanding and evaluating blind deconvolution algorithms,''
\newblock in {\em Proceedings of CVPR}, 2009, pp. 1964--1971.

\bibitem{you1996regularization}
Y.-L. You and M.~Kaveh,
\newblock ``A regularization approach to joint blur identification and image
  restoration,''
\newblock {\em IEEE Transactions on Image Processing}, vol. 5, no. 3, pp.
  416--428, 1996.

\bibitem{chan1998total}
T.~F. Chan and C.-K. Wong,
\newblock ``Total variation blind deconvolution,''
\newblock {\em IEEE transactions on Image Processing}, vol. 7, no. 3, pp.
  370--375, 1998.

\bibitem{nah2021ntire}
S.~Nah, S.~Son, S.~Lee, R.~Timofte, and K.~M. Lee,
\newblock ``Ntire 2021 challenge on image deblurring,''
\newblock in {\em Proceedings of CVPR}, 2021, pp. 149--165.

\bibitem{xu2014deep}
L.~Xu, J.~SJ Ren, C.~Liu, and J.~Jia,
\newblock ``Deep convolutional neural network for image deconvolution,''
\newblock in {\em Proceedings of NIPS}, 2014, pp. 1790--1798.

\bibitem{tao2018scale}
X.~Tao, H.~Gao, X.~Shen, J.~Wang, and J.~Jia,
\newblock ``Scale-recurrent network for deep image deblurring,''
\newblock in {\em Proceedings of CVPR}, 2018, pp. 8174--8182.

\bibitem{kupyn2019deblurgan}
O.~Kupyn, T.~Martyniuk, J.~Wu, and Z.~Wang,
\newblock ``Deblurgan-v2: Deblurring (orders-of-magnitude) faster and better,''
\newblock in {\em Proceedings of ICCV}, 2019, pp. 8878--8887.

\bibitem{9619959}
H.~Ma, S.~Liu, Q.~Liao, J.~Zhang, and J.-H. Xue,
\newblock ``Defocus image deblurring network with defocus map estimation as
  auxiliary task,''
\newblock {\em IEEE Transactions on Image Processing}, vol. 31, pp. 216--226,
  2022.

\bibitem{TIPpaper20161}
L.~D'Andr\`{e}s, J.~Salvador, A.~Kochale, and S.~S\"{u}sstrunk,
\newblock ``Non-parametric blur map regression for depth of field extension,''
\newblock {\em IEEE Transactions on Image Processing}, vol. 25, no. 4, pp.
  1660--1673, April 2016.

\bibitem{DefocusPaper}
S.~Zhuo and T.~Sim,
\newblock ``Defocus map estimation from a single image,''
\newblock {\em Pattern Recognition}, vol. 44, no. 9, pp. 1852 -- 1858, 2011.

\bibitem{karaali2017edge}
A.~Karaali and C.~R. Jung,
\newblock ``Edge-based defocus blur estimation with adaptive scale selection,''
\newblock {\em IEEE Transactions on Image Processing}, vol. 27, no. 3, pp.
  1126--1137, 2017.

\bibitem{pattern2020}
S.~Liu, Q.~Liao, J.-H. Xue, and F.~Zhou,
\newblock ``Defocus map estimation from a single image using improved
  likelihood feature and edge-based basis,''
\newblock {\em Pattern Recognition}, vol. 107, pp. 107485, 2020.

\bibitem{domainadapt}
J.~Lee, S.~Lee, S.~Cho, and S.~Lee,
\newblock ``Deep defocus map estimation using domain adaptation,''
\newblock in {\em Proceedings of CVPR}, June 2019.

\bibitem{tip2022}
A.~Karaali, N.~Harte, and C.~R. Jung,
\newblock ``Deep multi-scale feature learning for defocus blur estimation,''
\newblock {\em IEEE Transactions on Image Processing}, vol. 31, pp. 1097--1106,
  2022.

\bibitem{geman1984stochastic}
S.~Geman and D.~Geman,
\newblock ``Stochastic relaxation, gibbs distributions, and the bayesian
  restoration of images,''
\newblock {\em IEEE Transactions on pattern analysis and machine intelligence},
  , no. 6, pp. 721--741, 1984.

\bibitem{levin2007image}
A.~Levin, R.~Fergus, F.~Durand, and W.~T. Freeman,
\newblock ``Image and depth from a conventional camera with a coded aperture,''
\newblock {\em ACM transactions on graphics}, vol. 26, no. 3, pp. 70--es, 2007.

\bibitem{dabov2008image}
K.~Dabov, A.~Foi, V.~Katkovnik, and K.~Egiazarian,
\newblock ``Image restoration by sparse 3d transform-domain collaborative
  filtering,''
\newblock in {\em Proceedings of Image Processing: Algorithms and Systems VI},
  2008, vol. 6812, p. 681207.

\bibitem{graham2019blind}
L.~Graham and Y.~Yitzhaky,
\newblock ``Blind restoration of space-variant gaussian-like blurred images
  using regional psfs,''
\newblock {\em Signal, Image and Video Processing}, vol. 13, no. 4, pp.
  711--717, 2019.

\bibitem{unetpaper}
O.~Ronneberger, P.~Fischer, and T.~Brox,
\newblock ``{U-Net}: Convolutional networks for biomedical image
  segmentation,''
\newblock in {\em Proceedings of MICCAI}, 2015, pp. 234--241.

\bibitem{LaplacianM}
A.~Levin, D.~Lischinski, and Y.~Weiss,
\newblock ``A closed-form solution to natural image matting,''
\newblock {\em IEEE Transactions on Pattern Analysis and Machine Intelligence},
  vol. 30, no. 2, pp. 228--242, 2008.

\bibitem{Gastal}
E.~S.~L. Gastal and M.~M. Oliveira,
\newblock ``Domain transform for edge-aware image and video processing,''
\newblock {\em ACM TOG}, vol. 30, no. 4, pp. 69:1--69:12, 2011,
\newblock Proceedings of SIGGRAPH.

\bibitem{comb1}
D.~Krishnan and R.~Fergus,
\newblock ``Fast image deconvolution using hyper-laplacian priors,''
\newblock in {\em Proceedings of NIPS}, pp. 1033--1041. 2009.

\bibitem{comb2}
A.~Levin, R.~Fergus, F.~Durand, and W.~T. Freeman,
\newblock ``Image and depth from a conventional camera with a coded aperture,''
\newblock {\em ACM Transactions on Graphics}, vol. 26, no. 3, pp. 70–es, July
  2007.

\bibitem{zhang2016spatially}
X.~Zhang, R.~Wang, X.~Jiang, W.~Wang, and W.~Gao,
\newblock ``Spatially variant defocus blur map estimation and deblurring from a
  single image,''
\newblock {\em Journal of Visual Communication and Image Representation}, vol.
  35, pp. 257--264, 2016.

\end{thebibliography}

\end{document}